# Open source disease analysis system of cactus by artificial intelligence and image processing

Kanlayanee Kaweesinsakul

Department of Information Technology, Faculty of Information Technology and Digital Innovation,
King Mongkut's University of Technology North Bangkok, kanlayaneekaw95@gmail.com

Siranee Nuchitprasitchai

Department of Information Technology, Faculty of Information Technology and Digital Innovation,
King Mongkut's University of Technology North Bangkok, siranee.n@itd.kmutnb.ac.th

Joshua M. Pearce

Department of Materials Science & Engineering and Department of Electrical & Computer Engineering,
Michigan Technological University, Houghton, MI, US.
Équipe de Recherche sur les Processus Innovatifs (ERPI), Université de Lorraine, France,
pearce@mtu.edu

There is a growing interest in cactus cultivation because of numerous cacti uses from houseplants to food and medicinal applications. Various diseases impact the growth of cacti. To develop an automated model for the analysis of cactus disease and to be able to quickly treat and prevent damage to the cactus. The Faster R-CNN and YOLO algorithm technique were used to analyze cactus diseases automatically distributed into six groups: 1) anthracnose, 2) canker, 3) lack of care, 4) aphid, 5) rusts and 6) normal group. Based on the experimental results the YOLOv5 algorithm was found to be more effective at detecting and identifying cactus disease than the Faster R-CNN algorithm. Data training and testing with YOLOv5S model resulted in a precision of 89.7% and an accuracy (recall) of 98.5%, which is effective enough for further use in a number of applications in cactus cultivation. Overall the YOLOv5 algorithm had a test time per image of only 26 milliseconds. Therefore, the YOLOv5 algorithm was found to suitable for mobile applications and this model could be further developed into a program for analyzing cactus disease.

## 1 Introduction

Although historically considered a pest, there is a growing interest in cactus cultivation [1] as cacti do not require much space for growth, are easy to grow and can be grown in a wide variety of environmental conditions. Cactus have a wide number of uses [2] from an eco-friendly material for wastewater treatment [3], a source of food [4], a medicinal food [5] and for specific pharmacological uses [6], but the cactus itself is subject to many diseases [7]. Cactus disease can be a severe problem, that can often be treated if diagnosed promptly. Most of the diseases found in cactus are fungal diseases (e.g. anthracnose [8] is caused by the fungus colletotrichum) although canker disease and rot [9]. Plant rusts diseases are from an infection invades tissues until the stems wither and collapse [10]. There are also various insects that cause problems with cactus including mealybugs, aphids, and mollusks.

These problems for cactus growers can become worse if the infected cactus is untreated and the disease spreads to other plants. This research proposes a system for analyzing the symptoms of cactus disease using computer vision and artificial intelligence. Specifically, this study develops a model for the analysis of cactus disease using the Faster R-CNN [11] and YOLO (You Only Look Once) algorithm [12] technique and to analyze cactus disease automatically to be able to quickly heal and prevent damage to cactus. The diseases which are analyzed into six groups include 1) anthracnose group, 2) canker group, 3) group of disease caused by lack of care, 4) aphid disease group, 5) plant rusts disease group and 6) normal group. The YOLO enabled image processing for disease analysis is rated for effectiveness, precision and accuracy and the results are discussed.

## 2 Literature Review

Previous work has applied imaging to cacti. Carter and Leeuwen mapped cacti using digital aerial imaging [13]. Image analysis has also been used to identify plant diseases. For example, a decision support system for the diagnosis of longan leaf disease used a decision tree technique [14]. Such a system builds a knowledge base from the knowledge of expert advice. The content analysis consisted of leaf appearance, shape, color, characteristic, and disease of appearance. Using the C5.0 algorithm, the researchers were able to apply a decision tree diagnostic technique to their research using image analysis results and this technique to diagnose cacti [14]. Priya et al. are developing a system for plant disease monitoring to assist in the treatment of plant diseases [15]. In this case the primary goal is to prevent crop loss because in India 50% of the loss of crops is caused by plant diseases and the detection of the disease is difficult and requires a specialist to investigate [15]. The process used image acquisition, image conversion (convert to same size), and K-mean clustering and classification techniques along with temperature and humidity sensing [15]. Image analysis using a YOLO derivative MangoYolo was able to provide disease detection reports via SMS in real-time fruit counting systems on mango trees in Australia [16]. The results that provided individual tree yields of mangos from pictures of an entire tree of the MangoYOLO model were 98.3% accuracy at 15 ms, which is more than YOLO at 95.3% at 10 ms, and R-CNN at 95.0% at 37 ms [16]. Thus, the analysis of the image is most effective.

YOLO or You Only Look Once is a Real-time Object Detection Model that is notable for its speed and accuracy and can detect even objects that are nested [12]. There is a complex grid structure for each layer that is reduced in size in each layer. The principle is that in an image with several objects, YOLO tries to frame the objects and identify them with a basic model that has been trained. YOLO belongs to the single-shot object detector, unlike the RCNN family that does not have a separate network for Region Proposals (RPN) and depends on anchor points at different levels [17]. YOLO has been evolving v2 [18], v3 [19], v4 [20] and v5 [21].

The model architecture consists of three main components:
- Spine - A neural network that collects and reproduces image features on different image resolutions.
- Neck: A set of network layers that blend and combine image properties and transfer image properties to prediction layers.
- Header: predict image properties, create framing, border, and type predictions [22].

YOLO V5 has several models that have different structures, including V5s, V5m, V5l and V5x each of which takes different parameters and has different results and performance [22].

## 3 Research Methodology

In this study, the Faster R-CNN algorithm [23] is an extension of Fast R-CNN [11] used to address problems encountered by Fast R-CNN as it relies on a Selective Search algorithm that takes a long time to build regional offer. The selective search algorithm cannot be defined in the task of detecting specific objects and therefore may not be accurate enough to detect all target objects in the dataset. Faster R-CNN has been developed to provide superior performance to Fast R-CNN in terms of speed and accuracy in detecting objects. In addition, YOLOv5 algorithm was used because it provides higher accuracy and less time-consuming compared to YOLOv4 and YOLOv3 [24]. Specifically, YOLOV5s was selected because it is a smaller model, uses the least number of parameters, and the processing is the fastest compared to other models [21]. YOLO v5 is open source and is licensed under the GNU General Public License v3.0 [25]. The system of analysis of the disease of cactus by the YOLO algorithm is also open source and is located in the GitHub repository [https://github.com/Kanlayanee-k/cactus-doctor]. The following research steps were implemented.

### 3.1 Study the problems of cactus culture

From a review of cactus-growing websites and social media the problems of the cactus cultivation, it was found that the most important problems were cactus abnormality and disease. Since the conditions used for cultivation can have a profound impact on cultivation such as sunlight, moisture, species, and care of the cactus can damage the cactus and its growers; it is therefore imperative that it be treated promptly. Therefore, this study explores a method that allows cactus growers to identify cactus problems by themselves without having to wait for a specialist.

## 3.2 Prepare data for modeling

To create a model to recognize and learn the diseases of cacti, first images of each cactus disease are taken to model. And the information for creating the model is gathered in the attachment [https://docs.google.com/document/d/1-meFanhyB0QF9uwL63VppkETnKFaJNtTX7YHkpEFJok/edit?usp=sharing]. The researchers collected 225 images of cactus disease from various sources and applied image rotation techniques, which resulted in a total figure of 900 images. The images were divided into 6 classes, which are selected from the most common diseases in cactus in the literature and discussed in social media shown in [Figure 1](#) and described as follows:

- Anthracnose, which starts from a small black spot and then spreads into a blistering black tint on the skin of the cactus. There may be black latex coming out from the wound.
- Canker starts with a small dot. Later the wound is raised, turns brown, swells, and cracks into scabs, which are rough and hard. At the edge of the wound, there is a yellow, orange, or brown band.
- Lack of care classification starts with a range of visual cues including wrinkles or softness or scars or scratches or the color is not bright, or the skin is brownish-colored with small black spots which grow slowly.
- The aphid classification is made up of small insects that hide in areas that are hard to see. However, aphid damage causes absorption of water from various parts of the plant that causes the kink to wither. The growth is interrupted and may eventually kill the plant.
- The rusts classification is made up of plant rusts are yellow-orange spots like iron rust. The wound may be a protrusion and a crack. A speck of rust dust on the trunk tends to start from the base of the trunk.
- Finally, the non-disease group (normal) is made up of healthy cacti.

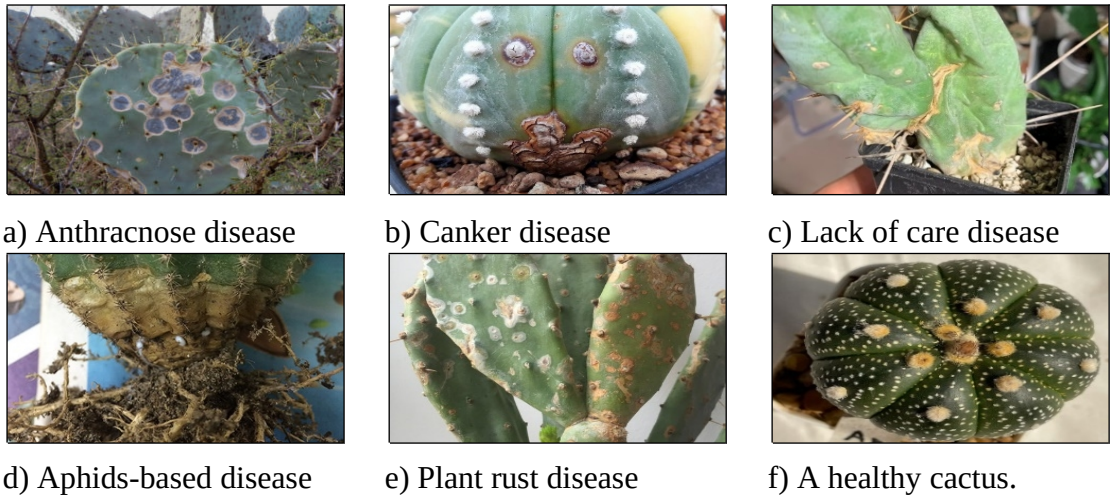

| a) Anthracnose disease | b) Canker disease | c) Lack of care disease |
| d) Aphids-based disease | e) Plant rust disease | f) A healthy cactus. |

Figure 1: **The six classes cactus analyzed.**

## 3.3 Create and test the model

### 3.3.1 Setup Environment.

A computer with an Intel Core i7-5820k and NVIDIA GeForce GTX970 4G OC series graphics processing unit (GPU) was used because the graphics processor improved the test of the model related to image processing analysis.

### 3.3.2 Setup Environment.

In the process of creating a dataset or dataset for testing and running a model, the data set used for the training data (Train) must be greater than the number of parameters. The smallest YOLOv5 structure contains 7.5 million parameters [21], with each set of equal or similar numbers to reduce the rate of bias. The Faster R-CNN, faster_rcnn_coco_inception_v2 was a chosen model [11]. To create a dataset it must be divided into three parts: Cactus_Training Dataset (60%), Cactus_Validation Dataset (20%), and Cactus_Test Dataset (20%). The number of samples for each data set for cactus disease is shown in Table 1.

Table 1: **Number of samples in each YOLO v5s set for cactus disease**

| Disease | Train | Test | Validation | Total |
|---|---|---|---|---|
| Anthracnose | 80 | 28 | 28 | 136 |
| Canker | 90 | 31 | 31 | 152 |
| Lack of care | 98 | 33 | 33 | 164 |
| Aphid | 84 | 28 | 28 | 140 |
| Plant rusts | 100 | 34 | 34 | 168 |
| Normal | 84 | 28 | 28 | 140 |
| Total | 536 | 182 | 182 | 900 |

Next, a directory is created for each dataset and label it to be stored by creating a structured directory as shown in Figure 2 into the directory named Training_data stored in yolov5 for the YOLOv5 algorithm and Faster_RCNN for the Fasrter R-CNN algorithm.

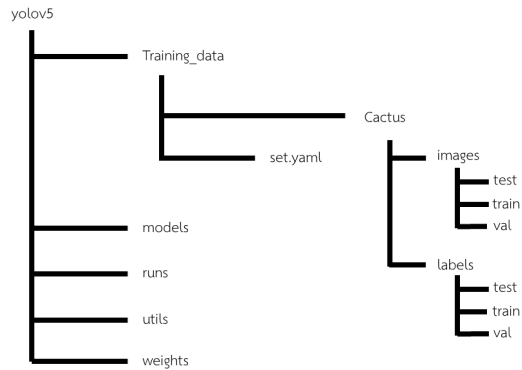

Figure 2: **yolov5 directory structure.**

Once the images have been created, the dataset is stored in the image directory in proportion to the number of samples following Table 1. After that, the data is labeled (Label) to be stored in the directory. The labels are files in the format [26] consisting of:
| class number of cactus disease | | cactus bounding box left X | | cactus bounding box top Y | | cactus bounding box right X | | cactus bounding box bottom Y |

The first value class number of cactus disease is the class of objects starting from 0 if there are 6 objects or 6 classes in the picture, such as anthracnose, canker, lack of care, aphid, normal, plant rusts then 0,1,2,3,4,5 and so on.

The second value cactus bounding box left X is the starting position of the X-axis of the object.

The third value cactus bounding box top Y is the starting position of the Y-axis of the object.

The fourth value cactus bounding box right X is the starting position of the X-axis of the object on the bottom right.

The fifth value cactus bounding box bottom Y is the starting position of the Y-axis of the bottom right object.

### 3.3.3 Training Model YOLO v5.

The modeling data training is the most time-consuming step in the process. The amount of time this will take depends on the size, number, and model used to test the data. In this system of analyzing the disease of the cacti, an S-sized YOLOv5 model was selected, which can select the structure of the model to train from the .yaml file extension stored in / yolov5 / models /. , which is the file for setting parameters, anchors, backbone and head. The faster_rcnn_coco_inception_v2 of Faster R-CNN model is chosen. Create a training file for the model data named set.yaml, which is stored in a folder named Training_data as a file for setting the data to import the training data. Each model was trained 600 epochs, which provided five disease classes and a normal class.

### 3.3.4 Model Tests for Cactus Disease Analysis.

In this research, an evaluation of a model with an accuracy of more than 90% is required to ensure that the model is performing sufficiently for practical applications, which Bramer concluded: "The performance of an acceptable model that is sufficient for use is 90% when the accuracy is set to class 0.5" [27]. Usually, there are four values to be considered when evaluating a model [28] which are:

1. True Positive_Cactus is the prediction that it is true and the answer (Ground Truth) tells you that there is real information that is predicted and correct compared to the answer.
2. True Negative_Cactus is a value that is predicted to be not real and the answer says it is not true.
3. False Positive_Cactus Is the expected value to be true But the answer says no.
4. False Negative_Cactus Is the value that is predicted to be nonexistent But the answer says it is true.

*Precision can thus be defined as:*

$$Precision_{(cactus)} = \frac{True\ Positive_{Cactus}}{True\ Positive_{Cactus} + False\ Positive_{Cactus}} = \frac{True\ Positive_{Cactus}}{n_{(cactus)}} \quad (1)$$

The prediction rate is against all predictions, where n is the number that the model attempts to predict.

*Recall is defined as:*

$$Recall_{(cactus)} = \frac{True\ Positive_{Cactus}}{True\ Positive_{Cactus} + False\ Negative_{Cactus}} \quad (2)$$

Recall is the predicted amount is equal to the total number of Ground Truths.
Training the model provides six figures of merit including for each epoch, which is the model training round including: , 1) Box loss, which is the number of frames that the model was trained not found, 2) Object loss, which is the number of objects that trained the model could not be found, and 3) Class loss is the number of detection classes found but incorrect results is provided. P, which is the precision is 4) and given by equation 1. In addition, 5) R is recall, which is defined by equation 2. Finally, 6 is mAP@.5 is mean precision when IoU is equal 0.5, thus, mAP.5:.95 is the average mAP in steps of 0.05 between 0.5 and 0.95 of IoU thresholds [22].

## 4   Research RESULTS

Data training and testing using the YOLOv5 S-sized model consisted of 6 classes and 600 training cycles. The results of the model tests are shown in Table 2.

Table 2: **Number of samples in each YOLO v5s set for cactus disease**

| Epoch | Box loss | Object loss | Class loss | P | R | mAP@.5 | mAP@.5:.95: |
|---|---|---|---|---|---|---|---|
| 100/599 | 0.02509 | 0.01376 | 0.014512 | 0.5552 | 0.8709 | 0.8082 | 0.5085 |
| 200/599 | 0.02144 | 0.01239 | 0.009312 | 0.6788 | 0.9459 | 0.9316 | 0.6521 |
| 286/599 | 0.01897 | 0.01082 | 0.008101 | 0.7668 | 0.9852 | 0.9634 | 0.7044 |
| 300/599 | 0.01959 | 0.01075 | 0.006623 | 0.7694 | 0.9653 | 0.9541 | 0.6846 |
| 400/599 | 0.01669 | 0.009881 | 0.005747 | 0.8469 | 0.9759 | 0.9638 | 0.7095 |
| 493/599 | 0.0147 | 0.009004 | 0.001714 | 0.8967 | 0.9758 | 0.9684 | 0.6997 |
| 500/599 | 0.01317 | 0.009029 | 0.003283 | 0.8852 | 0.9778 | 0.967 | 0.7039 |
| 531/599 | 0.01208 | 0.00829 | 0.002192 | 0.851 | 0.9747 | 0.9733 | 0.7171 |
| 533/599 | 0.01595 | 0.008852 | 0.002933 | 0.856 | 0.9805 | 0.9731 | 0.7214 |
| 599/599 | 0.01312 | 0.008298 | 0.003344 | 0.8456 | 0.9556 | 0.9653 | 0.7085 |

In this experiment, the precision was 89.67%, recall was 98.52%, and mean average precision is 97.33 %, as shown in [Figure 3.](#)

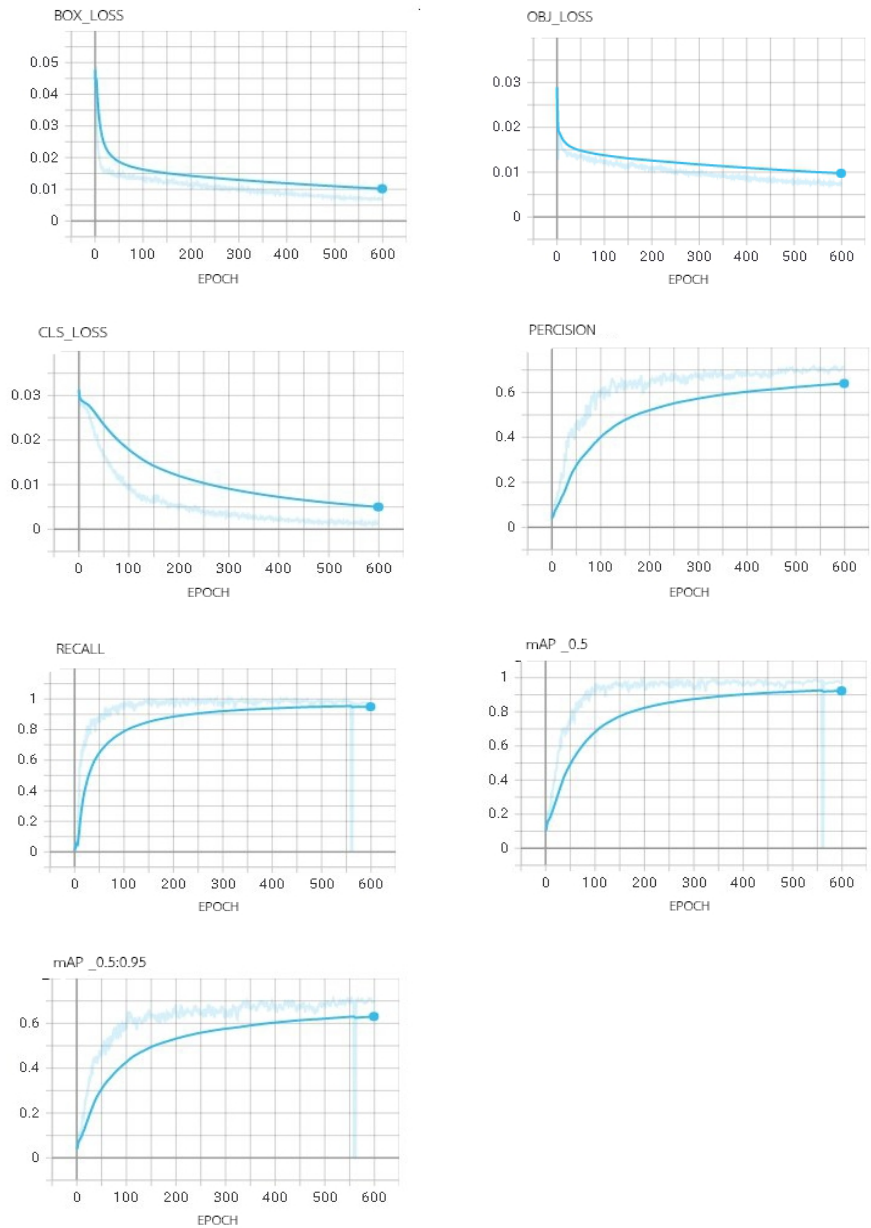

Figure 3: **The results of the test model.**

The performance results of Faster R-CNN and YOLOv5 (size s) models are presented in Table 3. The results show that the Faster R-CNN algorithm had a mAP (mean Average Precision) of 0.7008, loss of 0.1581, a training time of 1.5 hours, and a test time per image

of19 milliseconds. The YOLOv5 algorithm with mAP (mean Average Precision) of 0.9733, of 0.02042, training time of 50.9 hours, and test time per image of 26 milliseconds. Although the mean average preision of the Faster R-CNN algorithem was largely inferior to the YOLOv5 algorithm, he training time of the Faster R-CNN algorithm was one thirty-third of the YOLOv5 algorithm. The test time per image for both models was equivalent. Therefore, the YOLOv5 algorithm is more efficient and a better choice for this application, even though the Faster R-CNN algorithm is faster in training time.

Table 3: **Experimental output parameters of Faster R-CNN and YOLOv5 algorithms.**

| Parameter | Faster R-CNN | YOLOv5 (size s) |
| --- | --- | --- |
| mAP@.5 | 0.7008 | 0.9733 |
| Loss | 0.1581 | 0.02042 |
| Training Time | 1.5 Hours | 50.9 Hours |
| Test Time per image | 19 milliseconds | 26 milliseconds |

Results from the test model display the area where the image can be detected and to be processed with the possible model as seen in Figure 4.

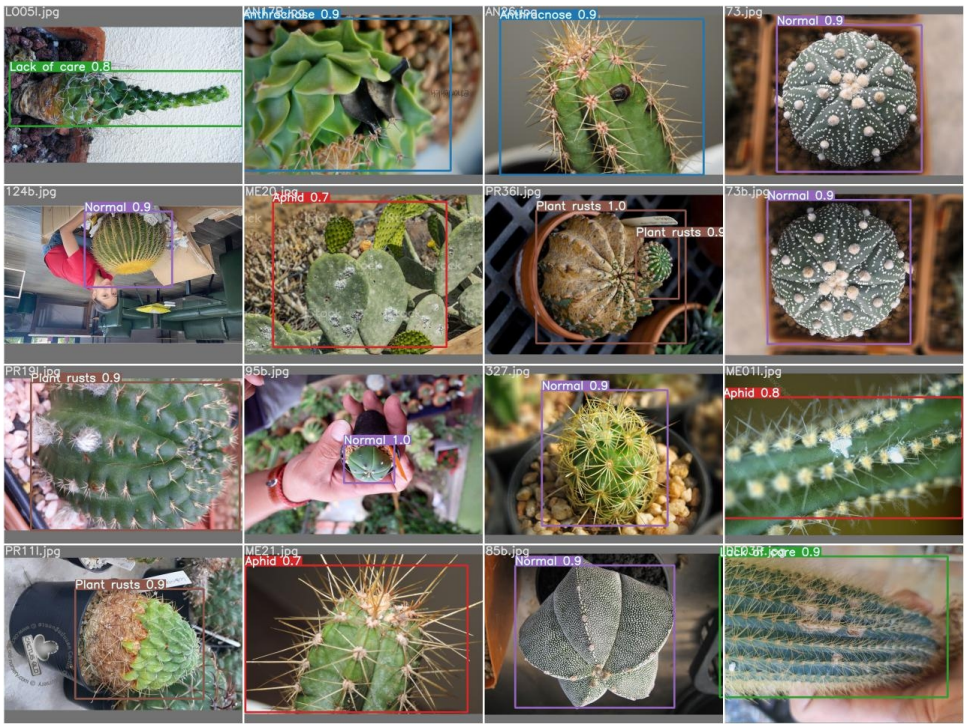

Figure 4: **The results of the test model.**

# 5 DISCUSSION AND FUTURE WORK

Based on the experimental results summarized in Table 3, the YOLOv5 algorithm was more effective at detecting and identifying cactus disease than the Faster R-CNN algorithm, and the test time per image is only 26 milliseconds. Therefore, the YOLOv5 algorithm was found to be suitable for mobile applications. As can be seen from the results an open source method of identifying cactus disease has been demonstrated effectively. This new application of YOLO can be adapted to be used by cactus owners to quickly diagnose disease on their smartphones and by commercial cactus farmers and retailers to automatically test for diseases and train new employees. Future work is needed to develop this smartphone app, as well as apply this algorithms technique to two broader applications. First, as many cacti species are threatened [29] this technique can be applied to remote monitoring of cacti (e.g. a drone with camera) for both conservation [30] as well as large-scale cactus farms for human food [31] or fodder [32] to quickly evaluate cactus disease. In addition, cactus plants are well known to survive in low-water environments, but some species can also survive cold temperatures and the winter [33], which makes them a potential source of alternative food for major global catastrophic risks that substantially reduce available rainfall [34,35]. Monitoring cactus health automatically using the method outlined here could help increase the number of available calories regardless of the state of the agricultural system.

# 6 Conclusions

This study presented a new model for the analysis of cactus disease using the YOLO algorithm technique that can identify the symptoms of cactus diseases by means of classification and image processing. The developed model was able to analyze 6 diseases of the cactus from the symptoms of the cactus: 1) anthracnose group, 2) canker disease group, 3) group of disease caused by lack of care, 4) aphid disease group, 5) plant rusts disease group and 6) normal group. The experimental results to identify cactus diseases with the Faster R-CNN and the YOLOv5 algorithms showedthat the YOLOv5 algorithm is far more accurate aat cactus disease detection and almost eight times lower in loss than the Faster R-CNN algorithm. The technique used trained data and was tested with S-size model for the 6 classes using 900 images in 600 cycles. The precision obtained was 0.8967 and the accuracy (recall) is 0.9852, which is effective enough for further use in a number of applications. It is clear that the open source tool developed and tested in this study if applied on mobile or web applications, will help those interested in growing cactus to identify the symptoms of the illness and take appropriate care of the cactus themselves. This will help reduce the damage that will occur to the cacti and growers.